# Perceptual Tone Mapping Model for High Dynamic Range Imaging

**IMRAN MEHMOOD[1], XINYE SHI[1], M. USMAN KHAN [1] and MING RONNIER LUO[1,2]**
[1]State Key Laboratory of Modern Optical Instrumentation, Zhejiang University, Hangzhou, China
[2]Leeds Institute of Textiles and Colour, University of Leeds, Leeds, UK

Corresponding author: Ming Ronnier Luo (e-mail: m.r.luo@zju.edu.cn).

**ABSTRACT** One of the key challenges in tone mapping is to preserve the perceptual quality of high dynamic range (HDR) images when mapping them to standard dynamic range (SDR) displays. Traditional tone mapping operators (TMOs) compress the luminance of HDR images without considering the surround and display conditions emanating into suboptimal results. Current research addresses this challenge by incorporating perceptual color appearance attributes. In this work, we propose a TMO ($TMO_z$) that leverages CIECAM16 perceptual attributes, i.e., brightness, colorfulness, and hue. $TMO_z$ accounts for the effects of both the surround and the display conditions to achieve more optimal colorfulness reproduction. The perceptual brightness is compressed, and the perceptual color scales, i.e., colorfulness and hue are derived from HDR images by employing CIECAM16 color adaptation equations. A psychophysical experiment was conducted to automate the brightness compression parameter. The model employs fully automatic and adaptive approach, obviating the requirement for manual parameter selection. $TMO_z$ was evaluated in terms of contrast, colorfulness and overall image quality. The objective and subjective evaluation methods revealed that the proposed model outperformed the state-of-the-art TMOs.

## I. INTRODUCTION

The HDR representation offers an unlimited tonal range and prioritizes the preservation of fine details. It allows for a more immersive visual experience when viewing movies, photographs, playing computer games, or inspecting visualizations [1, 2]. However, most consumer display devices are not equipped to handle such rich visual content. It is where a tone mapping algorithm comes in, which adjusts the tonal range of HDR data to match the capabilities of the device before displaying [3]. While this reduction in the tonal range can lead to a decrease in the original quality of HDR content as well [4].

Many TMOs have been proposed over the years [5-12]. The TMOs are based on image processing techniques, HVS, or sigmoidal functions. The recent evaluations of the TMOs report that algorithms perform differently in different scenarios [13, 14]. TMOs aim to produce an SDR image that perceptually matches the original HDR scene as closely as possible while selectively preserving significant visual features. However, traditional TMOs are often parametric and rely on subjective visual assessments to yield plausible results, leading to various artifacts, such as over-enhancement, over-stylization, halo effect, and blurring. These artifacts are particularly pronounced when the TMO parameters are not properly tuned for a given HDR scene. As a result, there is a growing need for scene-adaptive TMOs that can generate high-quality SDR images under diverse HDR scenes.

The HVS has the ability to adapt to both night and daytime, relying on two types of photoreceptor cells known as cones and rods [15]. The range of luminance that can only be perceived by the rods in the eye is known as scotopic vision. Photopic vision is the luminance range when cones are active [16]. The cones are more active in bright luminance levels and provide color vision. The rods and cones operate in a complementary fashion across a range of luminance levels. In addition, there is an intermediate range of luminance levels where both photoreceptor types contribute to the vision, referred as mesopic vision.

Khan *et al.* [17] proposed a histogram-based tone mapping technique and combined the perceptual quantizer (PQ) function. Initially, the luminance of the HDR image was transformed using PQ curve to compress the dynamic range.

Further, the algorithm constructed a histogram of the luminance values from the PQ-transformed data. It then limited pixel counts in the histogram to a predefined threshold determined using a uniform distribution model. This technique may not be suitable for all types of HDR images or scenes, as the effectiveness of the sensitivity model depends on the image content and the viewing conditions. The histogram-based approach may not be able to compress the full dynamic range of the HDR image, resulting in some loss of highlight or shadow detail.

Hui *et al.* proposed the clustering-based content and color adaptive tone mapping algorithm [18]. It used a clustering-based approach to segment the image into different regions based on color and content similarity. The algorithm transformed each color patch into three components: patch mean, color variation, and color structure. It then grouped these patches into different clusters. An adaptive subspace was learned for each cluster using principal component analysis (PCA). It allowed the patches to be transformed into a compact domain, facilitating more effective tone mapping.

The CNN-based TMOs, such as DeepTMO and TMO-NET, use convolutional neural networks (CNNs) to perform tone mapping [19, 20]. They generally comprise two main components: a CNN-based model and a post-processing stage. The CNN-based model is trained using a dataset of HDR and SDR image pairs to learn a mapping function that can convert an HDR image to an SDR image with visually pleasing results. The post-processing stage of a CNN-based TMO is used to enhance the quality of the output image produced by the CNN model.

Another group of methods is based on color appearance models (CAM). CAMs are models of color vision that predict color appearance under various viewing conditions. CIE proposed CIECAM02 [21] which was later superseded by CIECAM16 [22, 23] to rectify some mathematical problems and to simplify the model. The input parameters to the model include viewing conditions of the surround, luminance, color temperature, and background, and the output parameters are color appearance attributes including brightness, lightness, colorfulness, chroma, saturation, hue angle and hue compositions. The model accurately predicts various visual effects such as the effects of Hunt, Stevens, chromatic adaptation, and surround. The model has been applied in diverse fields, including color reproduction and image processing. Kuang and Fairchild developed iCAM06 [24], which is based on CIECAM02, and it is particularly useful for image applications aimed at reducing the dynamic range of an image, following the cones and rods response of the HVS. However, this successive sequence causes color distortions originating from the balance changes of the three color channels (XYZ) processed separately.

Li *et al.* enhanced iCAM06 by adding a multi-scale enhancement algorithm to correct the chroma of images by compensating for saturation and hue drift [5]. However, the compensated chroma and hue factors were depended on two parameters to be tuned per image. Moreover, the contrast compression used is the same as that of iCAM06. Hence, the contrast quality was the same as that of iCAM06.

Kwon *et al.* proposed [6] multi-layer decomposition-based tone mapping by modifying the iCAM06, similar to Li TMO. It repairs color distortion by applying a color difference map to the strong edges, utilizing iCAM06 contrast compression. The algorithm first decomposed the image into base and detail layers. The color distortion map was calculated by the difference between tone mapped XYZ image and tone mapped base layer. Since the difference was based on the tone mapped XYZ image, hence the color distortion map inherently contains distortions.

The previously proposed TMOs use either the luminance channel of the HDR image or compress each X, Y and Z coordinate. There is no work on compression of the brightness in HDR imaging. It is acquired that little work has been done in utilizing the perceptual properties of CAMs in tone mapping and color processing. By engaging the HVS phenomena, CEICAM16 can accurately model human perception of colors, making them useful for color-critical applications.

In this paper, a TMO is proposed utilizing perceptual brightness and image-appearance mapping based on perceptual colorfulness and hue. The tone mapping model is proposed, including brightness compression after the separation of brightness into base and detail layers. Next, the local contrast is enhanced and colorfulness and hue computation are derived. In the later sections, the brightness compression parameter is automated by a psychophysical experiment. The visual results are presented and the model is evaluated using objective and subjective methods.

## II. PROPOSED MODEL

The flowchart of the $TMO_z$ is illustrated in Figure 1. The input to the $TMO_z$ is XYZ HDR data acquired from an HDR image. The RGB to XYZ transformation can be accomplished using a camera characterization model. However, the manufacturer usually does not incorporate the camera characterization model in the image file properties. The characterization model can be developed in this case, or a color space conversion matrix can be employed without non-linearizing the gamma mapping equations. For instance, the HDR radiance map $(R_H, G_H, B_H)$ can be converted into XYZ using the sRGB transformation matrix [25].

$$\begin{pmatrix} X_H \\ Y_H \\ Z_H \end{pmatrix} = M_{sRGB} \begin{pmatrix} R_H \\ G_H \\ B_H \end{pmatrix} \qquad (1)$$

The luminance $Y_H$ is essential for the prediction of multiple luminance-dependent phenomena, such as adaptation luminance $L_a$ and key of the image.

The perceptual attributes such as brightness, lightness, hue and colorfulness of the HDR image can be calculated using a CAM. In $TMO_z$, CIECAM16 was used for the reasons explained previously. It requires the surround conditions such as background luminance $Y_b$ and the adaptation luminance $L_a$

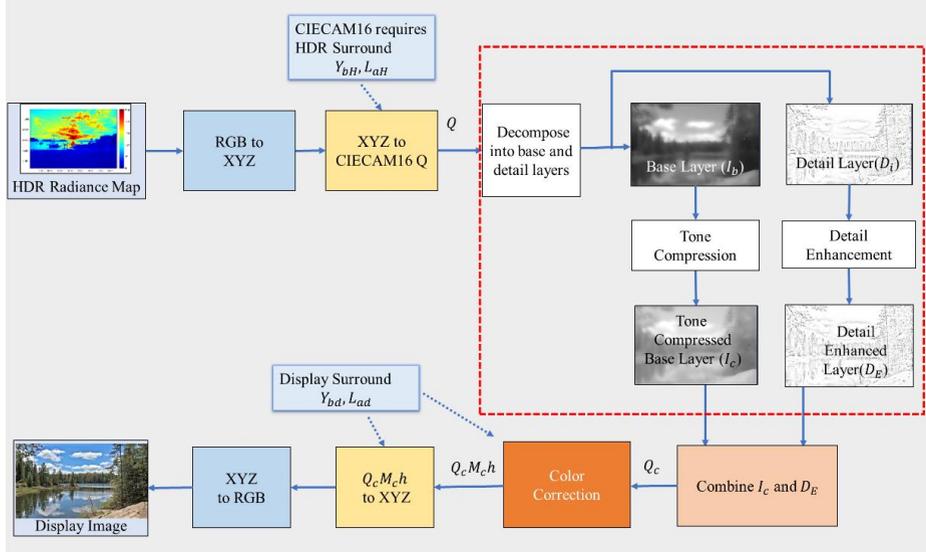

**FIGURE 1.** The flow chart of the proposed tone mapping model.

$L_a = L_w Y_b / 100$ are two major factors required while predicting such appearance attributes.

There are two types of surround conditions to be used in the algorithm, i.e., 1) to compute the brightness of the HDR image; the adaptation luminance $L_{aH}$, background luminance $Y_{bH}$ and, white point $X_{wH}, Y_{wH}, Z_{wH}$ and 2) to compute the hue and colorfulness from the HDR image under display conditions; the adaptation luminance $L_{ad}$, background luminance $Y_{bd}$ and white point $X_{wd}, Y_{wd}, Z_{wd}$.

### A. CIECAM16 BRIGHTNESS CALCULATION

The algorithm starts with calculations of standard CIECAM16 parameters. Firstly, calculate the input parameters using HDR white point $(X_{wH}, Y_{wH}, Z_{wH})$ as follows.

$$\begin{pmatrix} R_{wH} \\ G_{wH} \\ B_{wH} \end{pmatrix} = M_{CAT16} \begin{pmatrix} X_{wH} \\ Y_{wH} \\ Z_{wH} \end{pmatrix} \quad (2)$$

where $M_{CAT16}$ is the CAT16 chromatic adaption matrix. Calculate the degree of adaptation $D_H$.

$$D_H = F \left[1 - \left(\frac{1}{3.6}\right)\right] exp\left(\frac{-L_{AH} - 42}{92}\right) \quad (3)$$

Calculate other parameters as follows.

$$D_{RH} = D_H \frac{Y_{wH}}{R_{wH}} + 1 - D_H \quad (4)$$

Similarly, calculate $D_{BH}$ and $D_{GH}$ by replacing $R_{wH}$ with $B_{wH}$ and $G_{wH}$, respectively. Calculate,

$$F_{LH} = 0.2k^4(5L_{AH}) + 0.1(1 - k^4)^2(5L_{AH})^{1/3} \quad (5)$$

where

$$k = \frac{1}{5L_{AH} + 1}$$

and $n = \frac{Y_{bH}}{Y_H}, z = 1.48 + \sqrt{n}, N_{bb} = 0.725\left(\frac{1}{n}\right)^{0.2}$

$$N_{cb} = N_{bb}$$

$$\begin{pmatrix} R_{wcH} \\ G_{wcH} \\ B_{wcH} \end{pmatrix} = \begin{pmatrix} D_R R_{wH} \\ D_G G_{wH} \\ D_B B_{wH} \end{pmatrix} \quad (6)$$

$$R_{awH} = 400 \left(\frac{\left(\frac{F_{LH} R_{wcH}}{100}\right)^{0.42}}{\left(\frac{F_{LH} R_{wcH}}{100}\right)^{0.42} + 27.13}\right) + 0.1 \quad (7)$$

Similarly, calculate $G_{awH}$ and $B_{awH}$ by replacing $R_{awH}$ with $B_{awH}$ and $G_{awH}$, respectively. Calculate the achromatic response for the white point.

$$A_{wH} = [2R_{awH} + G_{awH} + 0.05B_{awH} - 0.305]N_{bb} \quad (8)$$

Now apply the formulation on HDR image XYZ $(X_H, Y_H, Z_H)$ as follows.

$$\begin{pmatrix} R_H \\ G_H \\ B_H \end{pmatrix} = M_{CAT16} \begin{pmatrix} X_H \\ Y_H \\ Z_H \end{pmatrix} \quad (9)$$

Calculate the color adaptation for the HDR image.

$$\begin{pmatrix} R_{cH} \\ G_{cH} \\ B_{cH} \end{pmatrix} = \begin{pmatrix} D_{RH} R_H \\ D_{GH} G_H \\ D_{BH} B_R \end{pmatrix} \quad (10)$$

Calculate the post-adaptation cone responses for HDR image.

$$R_{aH} = \begin{cases} -400 \left(\frac{(-x)^{0.42}}{(-x)^{0.42} + 27.13}\right) + 0.1, & R_{cH} < 0 \\ 400 \left(\frac{(x)^{0.42}}{(x)^{0.42} + 27.13}\right) + 0.1, & R_{cH} \geq 0 \end{cases} \quad (11)$$

where $x = F_{LH} R_{cH} / 100$. Similarly, $G_{aH}$ and $B_{aH}$ can be computed by replacing the $R_{cH}$ with $G_{cH}$ and $B_{cH}$, respectively. The chromatic response can be computed as

$$A_H = [2R_{aH} + G_{aH} + 0.05B_{aH} - 0.305]N_{bb} \quad (12)$$

and the lightness is given by (13).

$$J_H = 100 \left(\frac{A_H}{A_{wH}}\right)^{c.z} \quad (13)$$

The absolute predictor of the brightness of the HDR image is computed as

$$Q = (4/c)\sqrt{J_H/100} \times (A_{wH} + 4) F_{LH}^{0.25} \quad (14)$$

The brightness of the HDR image $Q$ is further decomposed and used for processing as in the following sections.

### B. Brightness Decomposition

The brightness $Q$ is decomposed into the base and detail layers to preserve the local contrast. Separating the brightness into two layers also reduces the halo artifacts. The tone compression is applied on the base layer while the detail layer is further enhanced to increase the local contrast of the resultant image.

The images can be decomposed into detail and base layers by subtracting a filtered image from the original image. The edge-preserving nonlinear bilateral filter is used to filter the $Q$ [26, 27]. In the Bilateral filter, each pixel is scaled by a Gaussian filter in the spatial domain and another Gaussian filter in the brightness domain. Hence the bilateral filter effectively preserves the sharp edges of the brightness image, preventing the halo artifacts in the images that are frequent in the local TMOs. The $Q$ is normalized with the maximum brightness $Q_{max}$ of the image and processed in log10 domain. The response of the bilateral filter to include normalized $Q$ can be written as in (15).

$$I_b = \frac{1}{k(s)} \sum_{p \in \Omega} f(p - s) g(Q_p - Q_s) Q_p \quad (15)$$

where $k(p)$ is the normalization factor for each pixel calculated by

$$k(s) = \sum_{p \in \Omega} f(p - s) g(Q_s - Q_p) \quad (16)$$

where $f(\cdot)$ and $g(\cdot)$ are spatial domain Gaussian filters and brightness domain filters with standard deviations $\sigma_s$ and $\sigma_r$, respectively. The $Q_s$ is the brightness at each pixel $s$. The bilateral filter controls the sharpness of the edges using standard deviations. The $\sigma_s$ is set to 2% of the image size while the value of the $\sigma_r$ is set to a constant value of 0.35 bilateral deviations. The $\sigma_s$ is set to 2% of the image size while the filter controls the sharpness of the edges using the standard value of the $\sigma_r$ is set to a constant value of 0.35.

The output of the bilateral filter is the base layer $I_b$ and it is subtracted from the normalized brightness image to extract the details layer $D_s$. Both layers are transformed back to the linear domain and combined after the following processing.

### C. BRIGHTNESS COMPRESSION

Typically, the TMOs compress the luminance channel and preserve the chromatic coordinates, resulting in over saturation of the tone mapped image. In TMO$_z$, the base layer of the brightness image is used for compression. Since the local contrast is enhanced (explained in the next section) using the detail layer, a simplified global tone compression method is used for global brightness compression. Moreover, the conversions in CIECAM16 are expensive; therefore, a simple global tone compression manages the processing time effectively. Hence, the low-frequency components of the base layer are compressed using a gamma function.

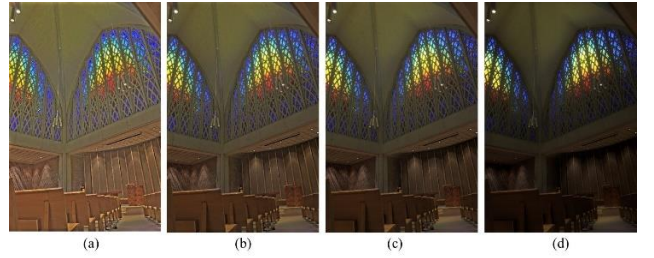

FIGURE 2. Variation in γ for the reproduction of tone mapped image using TMO$_z$. (a) $\gamma = 0.1$, (b) $\gamma = 0.4$, (c) $\gamma = 0.6$ and (d) $\gamma = 0.8$.

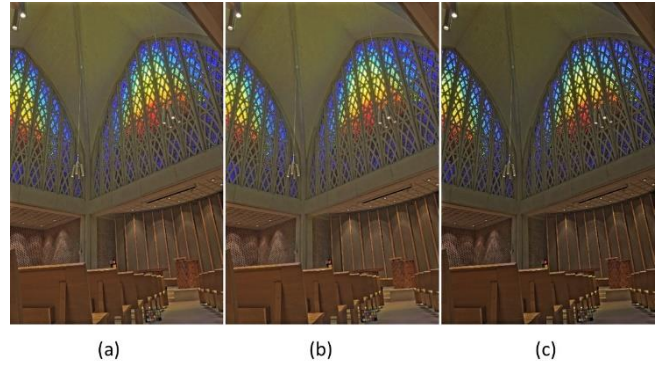

FIGURE 3. The effect of local contrast enhancement parameter. (a) β=0.8, (b) β=1.0 and (c) β=1.2.

$$I_c = A (I_b)^\gamma \quad (17)$$

where $A$ is the scaling factor and it was set to 1 in our model. The parameter $\gamma$ is the curve shaping parameter to compress the brightness. When the value of the gamma is increased, the compression is achieved higher along with the enhancement of the dark regions while the lower values of the gamma would decrease the effect of brightness compression. Figure 2 shows the images processed with different $\gamma$ values. The compression of brightness in turn results in the compression of contrast of the HDR image. The value of $\gamma$ is predicted automatically depending upon the key of the image. The modeling of the $\gamma$ is discussed later.

### D. LOCAL CONTRAST ENHANCEMENT

The separation of the base and detail layers preserves high-frequency information in the HDR image, preserving fine details. Since the base layer was compressed, the information in the low-frequency components was suppressed. The transformation of the tone mapped base layer and preserved detail layer does not contain adequate local contrast; therefore, the image's local contrast is enhanced using the following equation.

$$D_E = D_{i,max} \left(\frac{|D_i|}{D_{i,max}}\right)^\beta \times sign(D_i) \quad (18)$$

where, $D_{i,max}$ is the maximum value of $|D_i|$. The user-defined parameter $\beta$ controls the local contrast by stretching or reducing the high-frequency contents. The authors note that the value of the $\beta$ in the range of [1.1-1.3] produced the best results. In this work, the value of $\beta$ is kept at 1.1. The effect of local contrast enhancement is shown in Figures 3 (a)-(c). When the value of $\beta$ increases, the local contrast also increases. The $I_c$ and the $D_E$ are combined as $Q_c = Q_{\max}(I_c D_E)$ to get the final tone compression brightness.

### E. COLORFULNESS AND HUE CALCULATION

The hue for the tone mapped image can be computed directly from the HDR image and the same HDR surround conditions can be used. As discussed previously, human vision is influenced by viewing conditions. Specifically, the colorimetric information of the surround significantly influences the visual adaption mechanism. Hence, the colorfulness of the object depends on the illumination, background luminance, adaptation luminance and some other factors. Since, in CIECAM16, there is a crosstalk between colorfulness and brightness [28], the compression of the brightness of the HDR image ensues colorfulness distortions. Moreover, the brightness was normalized before separation into the base and details layer; hence, preserving simply the colorfulness of the HDR image for the tone-compressed brightness results in the image with low colorfulness (in contrast to the typical TMOs where the saturation increases when luminance is compressed) as depicted in Figure 4 (a). Hence, new colorfulness is required for the compressed brightness. The flow chart of the colorfulness calculation is depicted in Figure 5. The new colorfulness is HVS chromatic adapted using CIECAM16 under viewing conditions. The idea to calculate the colorfulness of the tone mapped brightness is by placing the tone mapped brightness in the display surround conditions however the color adaptation is carried from the HDR image using the same display viewing conditions. The colorfulness for the tone mapped image is calculated using the color adaptation equations from the HDR image, as follows.

#### 1) COMPUTE THE HUE

To compute the hue, calculate the post adaption responses $(R_{aH}, G_{aH}, B_{aH})$ of the HDR image calculated in the (10). The hue angle can be calculated as follows.

$$a_H = R_{aH} - \frac{12 \cdot G_{aH}}{11} + \frac{B_{aH}}{11}$$
$$b_H = \frac{R_{aH} + G_{aH} + 2 B_{aH}}{9} \quad (19)$$

$$h_H = tan^{-1}(b_H / a_H) \quad (20)$$

#### 2) COMPUTE THE COLORFULNESS

To calculate the colorfulness, first, calculate the achromatic response $A_{wd}$ and $F_{Ld}$ corresponding to the display white point $(X_{wd}, Y_{wd}, Z_{wd})$ using display surround conditions such as adaptation luminance $L_{Ad}$ and background luminance $Y_{bd}$ following (2)-(8). Now calculate the lightness of the tone mapped image using tone compressed brightness $I_c$ and $A_{wd}$ as follows.

$$J_c = 6.25 \cdot \left[ c \cdot \frac{I_c}{[A_{wd}+4] F_{Ld}^{0.25}} \right]^2 \quad (21)$$

Finally, the new colorfulness can be modeled as follows.

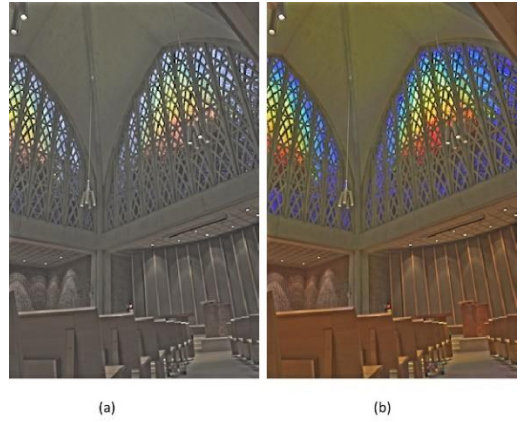

FIGURE 4. (a) The tone mapped image using the colorfulness of the original HDR image. (b) The image with the new colorfulness estimation.

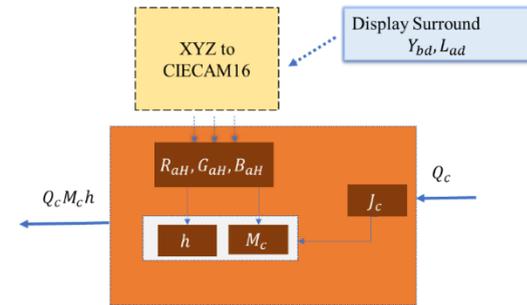

FIGURE 5. The flow chart for colorfulness and hue calculation.

$$e_t = \frac{1}{4} \cdot \left[ cos\left(\frac{h_H \cdot \pi}{180} + 2\right) + 3.8 \right] \quad (22)$$

$$t = \frac{\left(\frac{50000}{13}\right) \cdot N_c \cdot N_{cb}) \cdot e_t \cdot (a_H^2 + b_H^2)^{1/2}}{R_{aH} + G_{aH} + \left(\frac{21}{20}\right) \cdot B_{aH}} \quad (23)$$

$$C_c = t^{0.9} \left(\frac{J_c}{100}\right)^{0.5} (1.64 - 0.29^n)^{0.73} \quad (24)$$

$$M_c = C_c F_{Ld} \quad (25)$$

where the value of $c$, $N_c$ $N_{cb}$ are the same as in standard CIECAM16. The image with the new color estimation is depicted in Figure 4 (b). It can be seen that when the hue and colorfulness are calculated by employing the proposed method, the colors are restored in the tone mapped image.

### F. Display Image Transformation

Finally, the brightness, colorfulness, and hue image, i.e., $Q_C M_c h_H$ are transformed to XYZ coordinates using the inverse CIECAM16 model, as given [23]. For inverse transformation, display surround conditions are considered. To simulate glare, 1% pixels of the luminance values are clipped. The XYZ image is transformed to RGB images using XYZ to sRGB transformation or the specific display characterization model such as the GOG model. Note the images in the present paper are processed using sRGB signals.

### III. PARAMETER ESTIMATION

The brightness $Q$ of the HDR image was compressed using (17) with a control parameter $\gamma$. This parameter was modeled psychophysically to automate the TMO. The key of an image ($k$) is the statistical parameter that defines how bright the image is [29]. The key of the bright image will be relatively higher than the dark image. For example, a white-painted room will have a higher key than a dim object. The control parameter $\gamma$ can be written as

$$\gamma = f(k) \tag{26}$$

where $f(\cdot)$ is a function to be modeled psychophysically (see later at equation (30)) such that the $\gamma$ can be calculated automatically using the $k$, the key of the image, which can be calculated using equation (27).

$$k = 0.18 \times 4^{\frac{2\log_2 G_L - C_L}{C_L}} \tag{27}$$

where $G_L$ and $C_L$ are the image statistics referred to as geometric mean and contrast ratio, respectively, as given in equations (28) and (29).

$$C_L = \log_2 Y_{max} - \log_2 Y_{min} \tag{28}$$

$$G_L = \exp\left(\frac{1}{N}\sum \ln(\delta + Y_H)\right) \tag{29}$$

where $\delta$ is a small number to avoid logarithmic discontinuities and $Y_H$ is the luminance component of HDR image. The $Y_{min}$ and $Y_{max}$ are the minimum and maximum values of the $Y_H$.

## A. EXPERIMENT TO MODEL BRIGHTNESS CONTROL PARAMETER

A psychophysical experiment was conducted to model the brightness control parameter ($\gamma$). More than 300 HDR images were evaluated for their dynamic range and key values. Table I reports the minimum, maximum, and standard deviation (SD) of the key values for five databases, including RIT (the HDR photographic survey) [30], Stanford [31], MPI, and Funt [32, 33] databases. Thirty images were selected, spanning minimum to maximum key values (see $k$ value in equation (27)) from the RIT database since it contained the lowest and maximum key values among all the databases. The first image in the reading order in Figure 6 had the lowest key 0.085 and the last image had 0.8. These images included natural scenes, buildings, color checker charts, and many other objects captured at different times of the day.

Each image was tone mapped with the best values of the $\gamma$ estimated at the same display where the experiment had to be conducted. Later each image was rendered around the best values using $\gamma = \pm 0.1$ and $\pm 0.2$ and the best $\gamma$; resulting in five tone mapped images corresponding to one HDR image. Five rendered images are depicted in Figure 7. There were thirty HDR images therefore, the total rendered images were 150 (30 (HDR images) $\times$ 5 ($\gamma$ values)). Thirty percent of images were repeated to analyze the subject's variation and experiment validation; therefore, the total number of images for the experiment was 225. Typical viewing conditions were for CIECAM16 parameters to render the images i.e., luminance of reference white was set 560 cd/m² (equal to the peak luminance of the display) and background luminance was set to 0.2 cd/m² with average surround [34, 35].

Figure 8 shows the setup for the experiment. A pair of viewing distance of 1 meter was displayed side by side. The background of the images was set to black. The images were displayed on an Apple Pro Display XDR display located in a dark room and having a height of 41.2cm, width of 71.8cm and resolution of 2560 $\times$ 1440 pixels. The seating of the subjects was not fixed, and the sitting posture could be changed slightly to achieve comfort level. The peak luminance of the peak white of the display was set at CIE D65 and 1931 standard colorimetric observer at a luminance of 562 cd/m2.

TABLE I
THE STATISTICS OF THE KEY VALUES FOR VARIOUS DATABASES.

| Database | No. of Images | Maximum | Minimum | SD |
|---|---|---|---|---|
| RIT | 109 | 0.8093 | 0.0826 | 0.0566 |
| Stanford | 89 | 0.6930 | 0.0932 | 0.0379 |
| MPI | 9 | 0.7193 | 0.0823 | 0.0299 |
| Funt | 101 | 0.6993 | 0.0913 | 0.0469 |

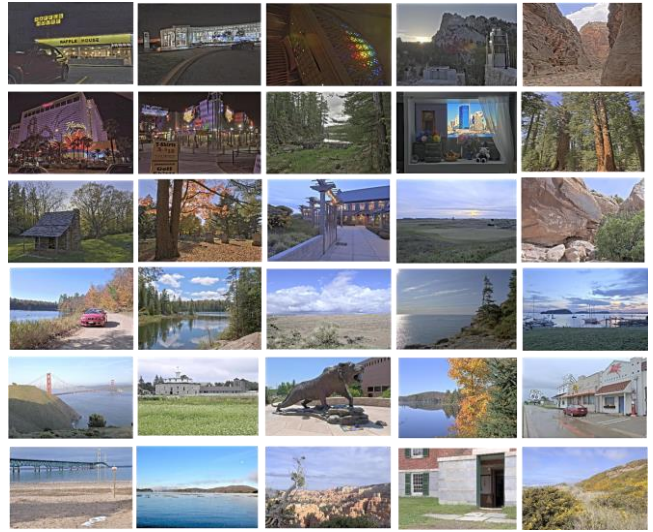

FIGURE 6. The images for the parameter estimation psychophysical experiment.

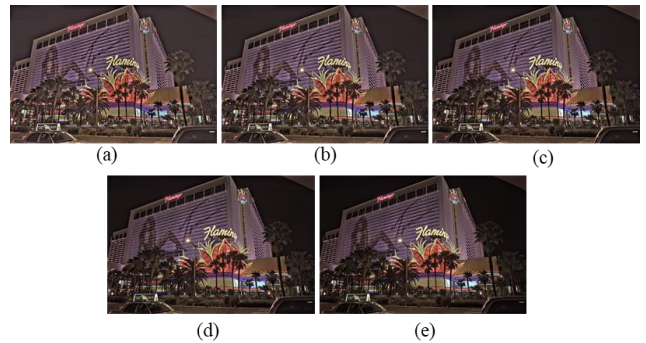

FIGURE 7. The rendered images using different values of gamma for the psychophysical experiment. (a) $\gamma = 0.18$, (b) $\gamma = 0.28$, (c) $\gamma = 0.38$, (d) $\gamma = 0.48$, (e) $\gamma = 0.58$.

For evaluation of spatial uniformity, the display was divided into 3 by 3 segments and the mean CEILAB color difference ($\Delta E_{ab}$) [36] calculated between the center and each segment was 1.1 $\Delta E_{ab}$. The gain-offset-gamma (GOG) display model was used for display characterization [37]. The 24 colors on the ColorChecker chart were used to check the rendered images having a field of view (FoV) of 22° at a $\Delta E_{ab}$ with a range of 1.13 $\Delta E_{ab}$ to 0.16 $\Delta E_{ab}$. The processed images were transformed to the display RGB using this model.

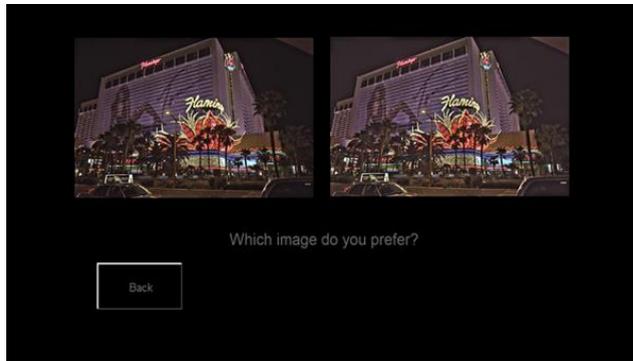

**FIGURE 8.** User interface for parameter estimation psychophysical experiment.

Ten subjects were recruited to perform the experiment. All the subjects were university students, and few were working in the color science domain. The mean age of the subjects was 25 with SD equal to 2.5. Before entering the experimental conditions, the subjects passed the Ishihara test [38] for the color vision test. A training session was performed for each subject to familiarize the interface and assessment scale. The subjects selected the preferred images assessing the overall image quality. Since there were 30 images and each image was rendered five times, each subject assessed 330 images (30 (images) $\times$ 5 $\times$ 4/2 = 300 + 10% repetitions for the subject's variation). Hence, the total assessments by ten subjects were 3,300 in the full experiment. Each subject engaged for 40 minutes on average to experiment.

### B. OBSERVER VARIATION
To analyze the validity of the experiment, inter- and intra-observer variations were calculated using the coefficient of variation (CV) [39]. It is a statistical measure used to express the degree of variation of a data set. It was computed as the ratio of the SD to the mean of the data, expressed in percentage. The CV is a dimensionless quantity often used to compare the variability of data sets with different units or scales. Higher values of CV indicate higher variability.

For inter-observer variability, the standard deviation was divided by the mean of the two values for the repeated images. The same pair of images shown to different observers were used to calculate intra-observer variability. A lower CV value of intra-observer implied better observer performance and a lower intra-observer variation inferred better reliability of the experiment. The mean value of inter-observer variability was 21, whereas the values for intra-observer variability were 18. Hence, the data was inferred as reliable for further analysis.

### C. MODELING THE PARAMETER
The raw data from the experiment was analyzed to model the curve shaping parameter $\gamma$. The percentage of preferred images ($P\%$) was calculated and $\gamma$ values at 50% were selected for further modeling. In a pair of two images preferred image over the other image in that pair. $P\%$ of 30 images are plotted in Figure 9. In each plot, the y-axis is the percentage of preference scores and the x-axis is the $\gamma$ value. The preference scores of each image were optimized and fitted using polynomial curve fitting. The scores of a few images were linearly fitted, while other images were fitted using 2-5 degree polynomials. The minimum Pearson's coefficient of determination $R^2$ was equal to 0.9 [40]. The experimental scores are plotted in red markers. The fitted curve is plotted using lines. The selected $\gamma$ values are pointed using green markers.

The $\gamma$ values at 50% were used to find a relationship between the key of the image ($k$ as discussed previously) and the $\gamma$. Linear regression was applied to the model given in the following equation.

$$\gamma = ak + b \qquad (30)$$

where the constants $a$ and $b$ optimized using regression are 0.6781 and 0.3128, respectively. Figure 10 shows the correlation between experimental values and model values. The $R^2$ was 0.82, which implied that the model performance agreed well with the psychophysical data. Hence, the brightness compression was made adaptive using this model.

## IV. EVALUATION OF THE TMO

### A. VISUAL COMPARISON
The performance of the $TMO_z$ was compared with ten state-of-the-art TMOs. Several TMOs were compared with $TMO_z$ such as Li [5] Kwon [6], Khan [17], Yang [41], Hui [18], Liang [42], Schlick [43], Durand [26], Kim [44], Drago [45], Lischinski [46], Meylan [47] and Reinhard [7]. Note only Schlick, Kim, Drago, and Reinhard TMOs were implemented by the authors and the others were taken from the websites of the original authors. Since the $TMO_z$ was fully adaptive, the suggested parameters for the other TMOs were used. Five example images are illustrated in Figures 11-14 to compare different TMOs' performance.

The tone mapped images in Figure 11 contain two Macbeth ColorChecker charts and an animal object placed in highlights and shadow areas indicated by green and red arrows, respectively. It is clear from the figure that the $TMO_z$ exhibited superior performance in comparison to other TMOs. $TMO_z$ produced a greater details and color information in shadows as well as in highlight regions. It is noted that human vision experiences low colorfulness in dark or dim lighting conditions. This effect is also prominently visible in the dark regions, particularly in the colors of the ColorChecker chart placed in the shadows.

Similarly, Figures 12-14 showed that the $TMO_z$ gave overall better preference among details, contrast, and colorfulness. The details in the highlights and shadows are well exposed, while the local contrast and colors are preserved for naturalness. While the other TMOs suffer from colorfulness or contrast distortions.

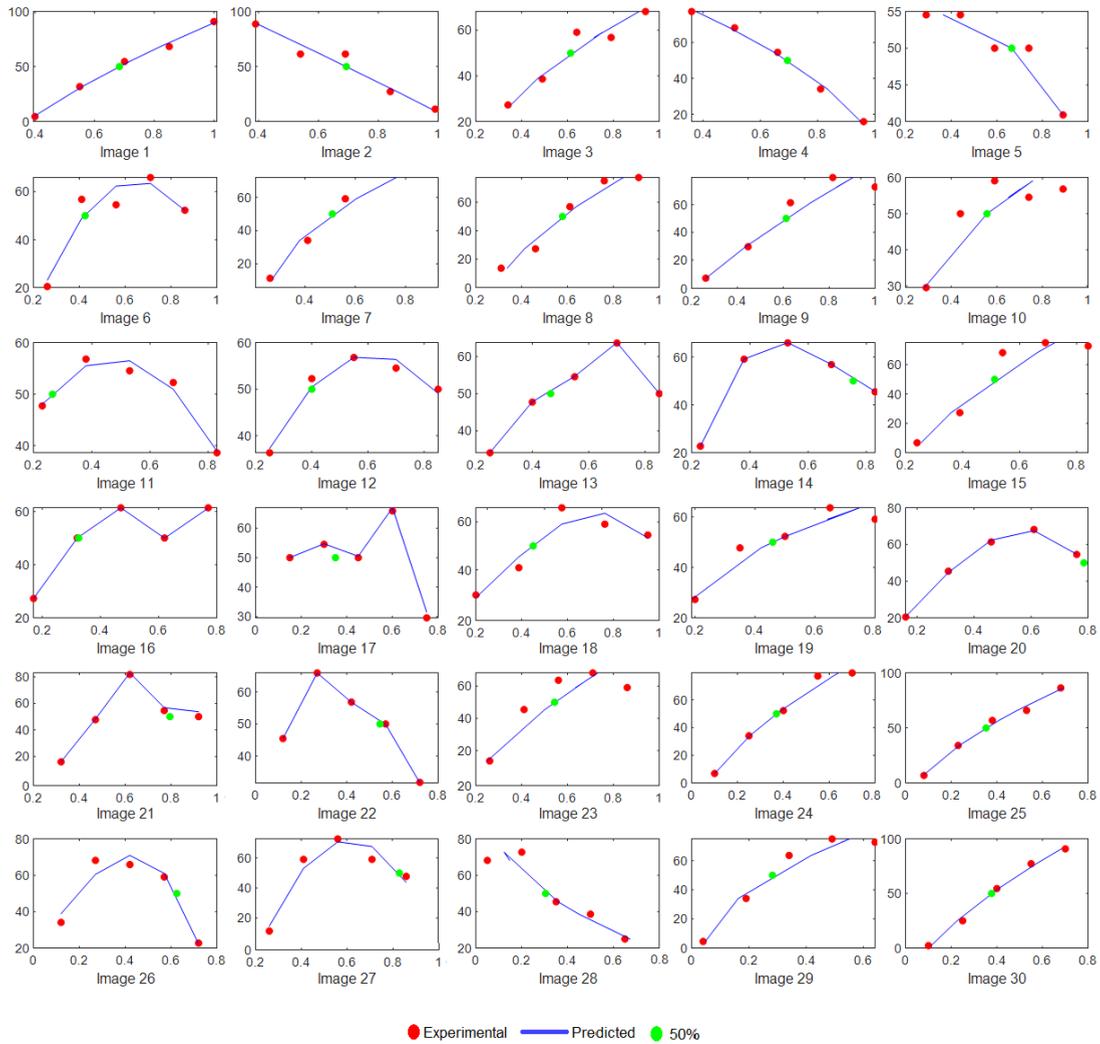

**FIGURE 9.** Percentage of preference scores and their prediction using polynomial modeling. The y-axis is the percentage of preference scores and the x-axis is the γ value.

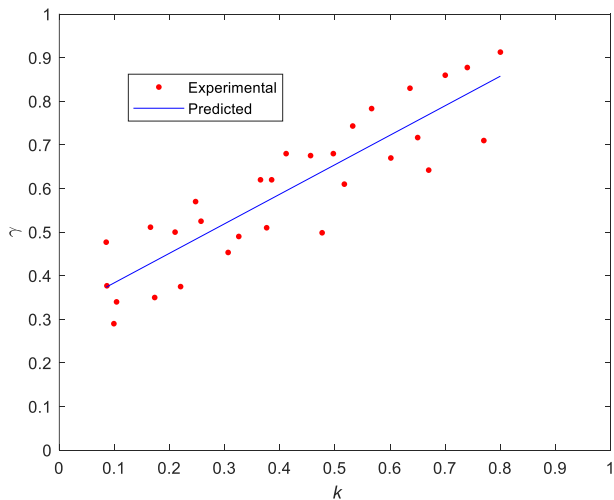

**FIGURE 10.** The correlation between experimental and prediction values.

### B. OBJECTIVE EVALUATION

For quantifying the performance of the TMOs, reduce reference image quality metrics (RR-IQMs) and no reference IQMs (NR-IQMs) were used (with/without an HDR reference image, respectively). The RR-IQMs utilize a subset of features from reference images to predict image quality on the reference image. The RR-IQMs utilize a subset of features from reference images to predict image quality while NR-IQMs predict the quality of the image without relying on the reference image. Two RR-IQMs, tone mapped image quality index (TMQI) [48] and FSITM-TMQI [49] were used. TMQI proposed by Yeganeh *et al.* [48] predicts the quality of tone-mapped images by combining a structural fidelity measure and a naturalness measure. It computes the structuredness, naturalness and overall quality of the images. The structural fidelity measure compares the test and reference image's structural similarities at different scales. In contrast, the naturalness measure models test image brightness and contrast using Gaussian and Beta probability density functions. Its advantage is that the structural similarity measure can compare tone mapped images with the HDR reference images.

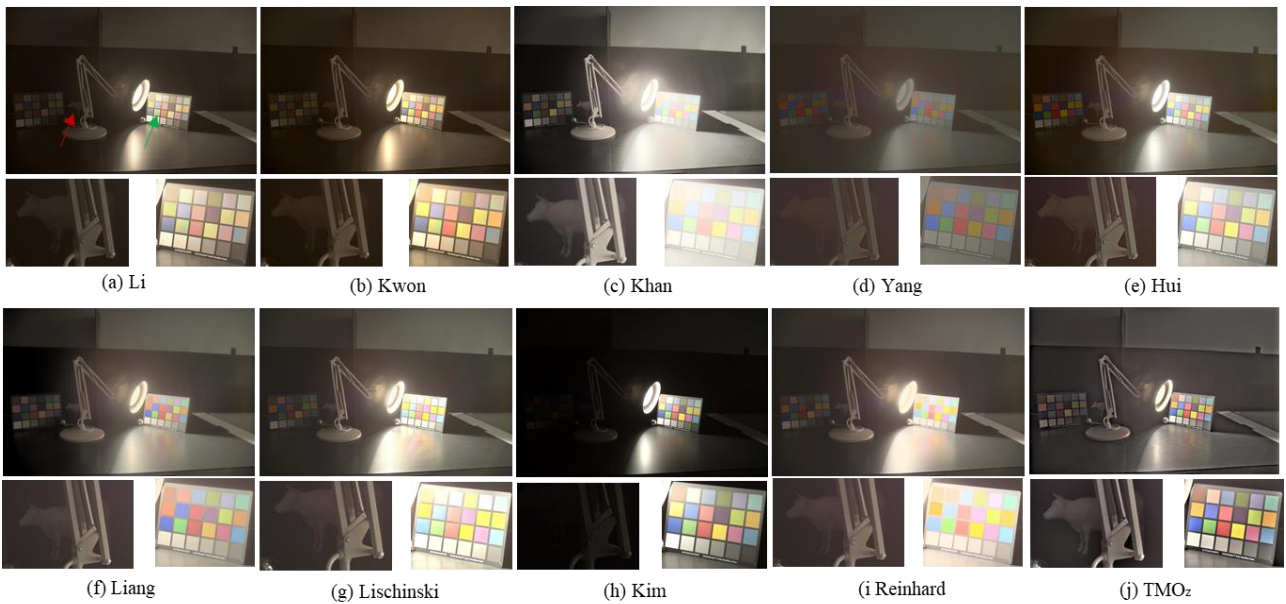

**FIGURE 11.** TMO$_z$ compared with the various TMOs. The ColorChecker chart reveals that TMO$_z$ is more preferred.

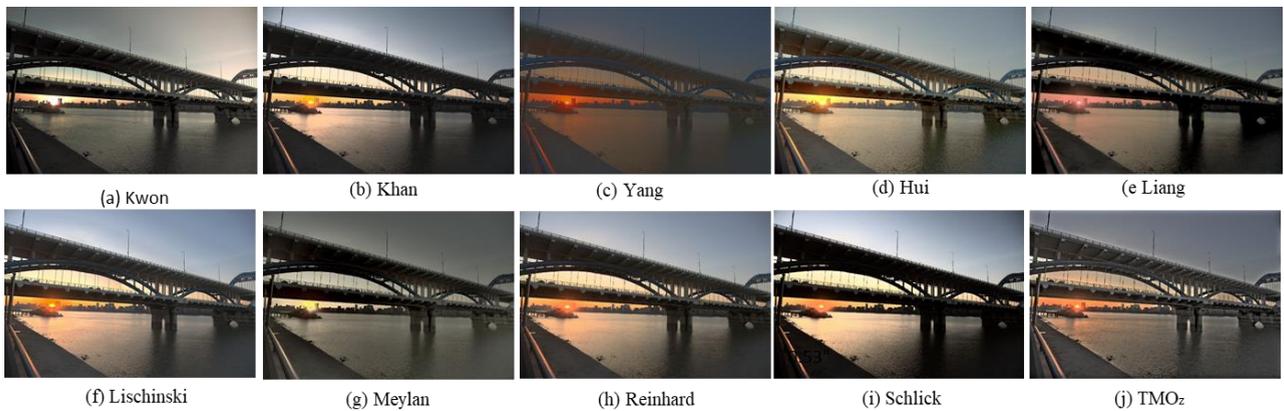

**FIGURE 12.** TMO$_z$ produced perfect colors around the sun while revealing the details in shadows and highlights better than other TMOs.

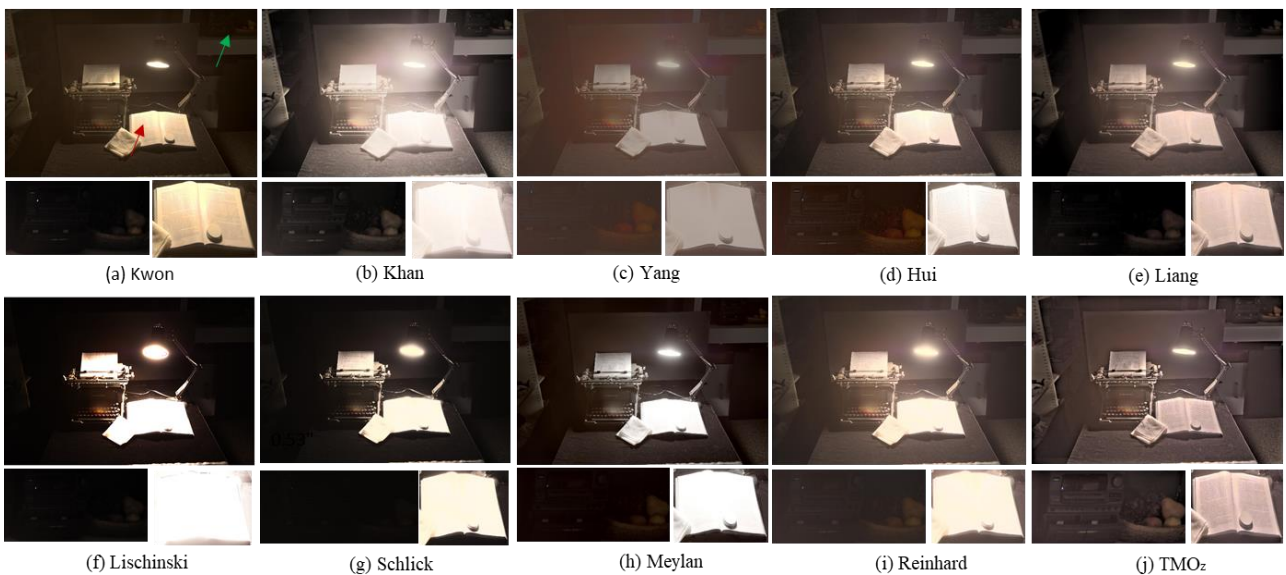

**FIGURE 13.** TMO$_z$ produces details in shadows and highlights better than other TMOs, as highlighted by the book and the corner areas.

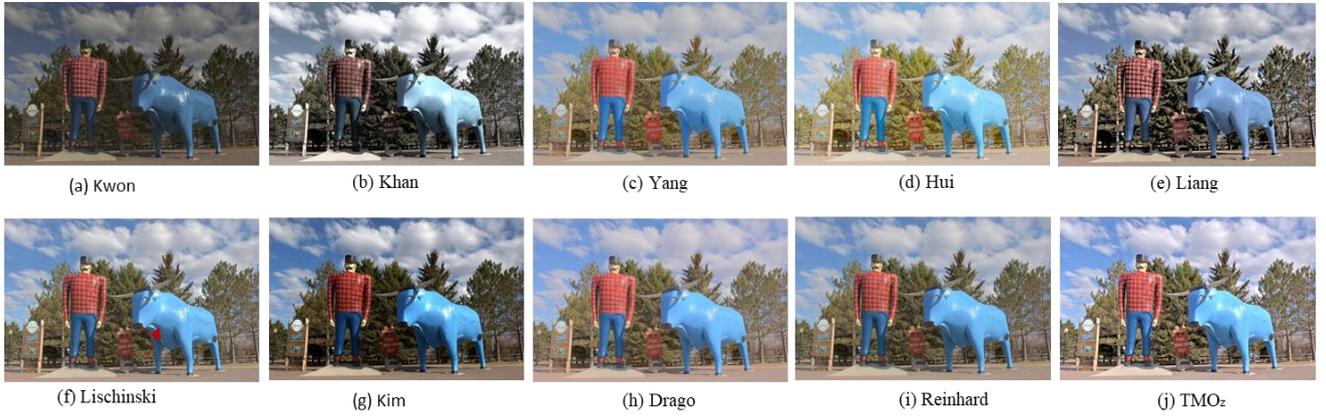

FIGURE 14. TMO$_z$ produced a more natural and brightest scene. The details in clouds are visible along with more accurate red, blue and green colors.

TABLE II
Comparison TMO$_z$ in terms TMQI structuredness and naturalness scores with other TMOs.

| | Structuredness (↑) | | | | | | | | | Naturalness (↑) | | | | | | | | |
|---|---|---|---|---|---|---|---|---|---|---|---|---|---|---|---|---|---|---|
| Image | Li | Kwon | Khan | Yang | Hui | Liang | Reinhard | Meylan | TMO$_z$ | Li | Kwon | Khan | Yang | Hui | Liang | Reinhard | Meylan | TMO$_z$ |
| Frontier | 0.76 | 0.78 | **0.89** | 0.69 | 0.61 | 0.70 | 0.58 | 0.62 | 0.82 | 0.23 | 0.24 | **0.80** | 0.01 | 0.27 | 0.12 | 0.19 | 0.23 | 0.60 |
| PepperMill | 0.87 | 0.89 | **0.89** | 0.76 | 0.82 | 0.85 | 0.77 | 0.84 | 0.87 | 0.29 | 0.29 | **0.78** | 0.02 | 0.46 | 0.05 | 0.32 | 0.25 | 0.72 |
| 507 | 0.90 | 0.91 | 0.82 | 0.73 | 0.82 | 0.87 | 0.87 | 0.88 | **0.92** | 0.43 | 0.44 | 0.32 | 0.16 | 0.50 | 0.58 | 0.37 | 0.54 | **0.86** |
| Pond | 0.85 | 0.87 | **0.86** | 0.75 | 0.72 | 0.79 | 0.76 | 0.76 | 0.94 | 0.53 | 0.54 | 0.76 | 0.47 | 0.57 | 0.96 | 0.54 | 0.40 | 0.47 |
| KitchenInside | 0.84 | 0.85 | 0.86 | 0.78 | 0.78 | 0.87 | 0.73 | 0.83 | **0.88** | 0.10 | 0.02 | 0.29 | 0.02 | 0.09 | 0.04 | 0.05 | 0.05 | **0.63** |
| DoubleChecker | 0.47 | 0.48 | 0.52 | 0.29 | 0.49 | 0.50 | 0.43 | 0.43 | **0.67** | 0.10 | 0.20 | 0.02 | 0.00 | 0.02 | 0.01 | 0.03 | 0.01 | **0.12** |
| RedwoodSunset | 0.78 | 0.79 | 0.64 | 0.54 | 0.62 | 0.64 | 0.64 | 0.66 | 0.92 | 0.61 | 0.62 | 0.49 | 0.14 | 0.65 | 0.45 | 0.13 | 0.10 | 0.48 |
| HooverGarage | 0.77 | 0.78 | 0.88 | 0.72 | 0.83 | 0.87 | 0.79 | 0.87 | **0.87** | 0.01 | 0.01 | 0.27 | 0.04 | 0.26 | 0.06 | 0.16 | 0.12 | **0.50** |
| BarHarborSunrise | 0.89 | 0.91 | 0.90 | 0.68 | 0.81 | 0.87 | 0.85 | 0.88 | **0.88** | 0.61 | 0.61 | 0.28 | 0.06 | 0.19 | 0.44 | 0.12 | 0.20 | **0.86** |
| BenJerrys | 0.90 | 0.91 | 0.91 | 0.77 | 0.85 | 0.89 | 0.92 | 0.91 | **0.96** | 0.82 | 0.83 | 0.79 | 0.21 | 0.08 | 0.97 | 0.57 | 0.33 | **0.80** |
| Mean | 0.80 | 0.82 | 0.82 | 0.67 | 0.74 | 0.79 | 0.73 | 0.77 | **0.87** | 0.37 | 0.38 | 0.48 | 0.11 | 0.31 | 0.37 | 0.25 | 0.22 | **0.60** |

TABLE III
Comparison TMO$_z$ in terms TMQI and FSITM-TMQI scores with other TMOs.

| | TMQI (↑) | | | | | | | | | FSITM-TMQI (↑) | | | | | | | | |
|---|---|---|---|---|---|---|---|---|---|---|---|---|---|---|---|---|---|---|
| Image | Li | Kwon | Khan | Yang | Hui | Liang | Reinhard | Meylan | TMO$_z$ | Li | Kwon | Khan | Yang | Hui | Liang | Reinhard | Meylan | TMO$_z$ |
| Frontier | 0.80 | 0.81 | **0.94** | 0.72 | 0.77 | 0.76 | 0.74 | 0.76 | 0.89 | 0.81 | 0.82 | 0.88 | 0.79 | 0.78 | 0.77 | 0.56 | 0.79 | **0.91** |
| PepperMill | 0.84 | 0.85 | **0.94** | 0.75 | 0.87 | 0.79 | 0.83 | 0.83 | 0.93 | 0.85 | 0.87 | 0.90 | 0.81 | 0.84 | 0.83 | 0.56 | 0.84 | **0.90** |
| 507 | 0.88 | 0.89 | 0.84 | 0.78 | 0.88 | 0.90 | 0.87 | 0.90 | **0.96** | 0.87 | 0.88 | 0.87 | 0.83 | 0.86 | 0.88 | 0.57 | 0.88 | **0.93** |
| Pond | 0.88 | 0.89 | **0.93** | 0.85 | 0.86 | 0.94 | 0.86 | 0.84 | 0.90 | 0.89 | 0.89 | **0.92** | 0.88 | 0.85 | 0.91 | 0.58 | 0.85 | 0.90 |
| KitchenInside | 0.76 | 0.78 | 0.85 | 0.76 | 0.78 | 0.79 | 0.75 | 0.78 | **0.91** | 0.81 | 0.82 | 0.85 | 0.83 | 0.82 | 0.85 | 0.58 | 0.84 | **0.91** |
| DoubleChecker | 0.63 | 0.64 | 0.67 | 0.55 | 0.66 | 0.65 | 0.63 | 0.62 | **0.76** | 0.74 | 0.75 | 0.73 | 0.73 | 0.75 | 0.76 | 0.56 | 0.74 | 0.74 |
| RedwoodSunset | 0.87 | 0.89 | 0.82 | 0.71 | 0.84 | 0.81 | 0.75 | 0.74 | 0.90 | 0.8 | 0.84 | 0.82 | 0.81 | 0.84 | 0.85 | 0.58 | 0.82 | **0.85** |
| HooverGarage | 0.74 | 0.75 | 0.85 | 0.74 | 0.83 | 0.79 | 0.80 | 0.81 | **0.88** | 0.79 | 0.80 | 0.86 | 0.81 | 0.83 | 0.83 | 0.57 | 0.84 | **0.88** |
| BarHarborSunrise | 0.90 | 0.92 | 0.86 | 0.74 | 0.81 | 0.88 | 0.81 | 0.83 | **0.95** | 0.89 | 0.90 | 0.87 | 0.82 | 0.83 | 0.89 | 0.58 | 0.86 | **0.92** |
| BenJerrys | 0.90 | 0.91 | 0.91 | 0.81 | 0.80 | 0.97 | 0.91 | 0.87 | **0.92** | 0.93 | 0.91 | 0.92 | 0.86 | 0.83 | 0.94 | 0.59 | 0.89 | **0.92** |
| Mean | 0.82 | 0.83 | 0.86 | 0.74 | 0.81 | 0.83 | 0.80 | 0.80 | **0.90** | 0.84 | 0.85 | 0.86 | 0.82 | 0.82 | 0.85 | 0.57 | 0.84 | **0.89** |

Feature similarity index for ton mapped images (FSITM) evaluates the performance of the TMOs by comparing the locally weighted mean phase angle map of the original HDR image to that of its corresponding tone-mapped image. The proposed index was found to outperform TMQI. Furthermore, a combination of the FSITM and TMQI named FSITM-TMQI produced even better results, demonstrating higher performance in evaluating the quality of tone-mapped images. TMO$_z$ was compared using TMQI structuredness, naturalness, and overall quality scores for ten images as depicted in Figure 15. The scores are reported in Tables II and III. The scores for each metric ranged from 0 to 1.0, with higher scores indicating better quality. The highest-scoring TMOs in each case were emphasized in bold text. It can be observed that TMO$_z$ performed best in 6 out of 10 images for naturalness, 7 out of 10 for structuredness, and overall image quality. Additionally, the average scores of TMO$_z$ were the highest for each metric.

The scores representing the performance of TMO$_z$, TMQI-FSITM are reported in Table III. It ranked highest for 9 images. The mean scores reported that TMO$_z$ performed best among all the state-of-the-art TMOs.

Table IV reports the average scores calculated for 104 images using TMQI and FSITM-TMQI. The TMO$_z$ outperformed the other metrics. Moreover, Rana *et al.* reported TMQI scores for 14 TMOs with maximum scores of 0.88 for the same dataset [19]. However, TMO$_z$ achieved 0.92, conveying that TMO$_z$ has better image quality compared to most of the TMOs available in the literature.

TABLE IV
Comparison of the TMO$_z$ with various TMOs in terms of R-IQMs and NR-IQMs using average scores of 104 images.

| TMO | R-IQMs | | NR-IQMs | | |
|---|---|---|---|---|---|
| | TMQI (↑) | FSITM-TMQI (↑) | PIQE (↓) | BRISQUE (↓) | BTMQI (↓) |
| Li | 0.86 | 0.87 | 36.86 | 22.98 | 4.00 |
| Kwon | 0.88 | 0.88 | 36.32 | 22.64 | 3.94 |
| Khan | 0.91 | 0.89 | 38.46 | 21.07 | 3.96 |
| Yang | 0.82 | 0.84 | 35.85 | 22.13 | 4.16 |
| Hui | 0.88 | 0.80 | 36.10 | 21.60 | 4.36 |
| Liang | 0.89 | 0.89 | 37.61 | 24.22 | 3.89 |
| Reinhard | 0.91 | 0.89 | 37.29 | 24.40 | 5.29 |
| Meylan | 0.88 | 0.89 | 37.86 | 25.05 | 4.34 |
| TMO$_z$ | **0.92** | **0.93** | **35.26** | **20.62** | **3.78** |

The TMO$_z$ was also evaluated using three NR-IQMs, namely PIQE, BRISQUE, and BTMQI. The PIQE method estimates the quality score of an image by analyzing its block-wise distortion, with the final score calculated as the mean of scores in the distorted blocks. PIQE uses a table to assign quality scale to the image based on its score range.

BRISQUE employs a machine learning model trained on a large dataset of high-quality images to estimate the perceived image quality. The algorithm extracts several statistical features from the image, such as local image contrast, luminance, and natural scene statistics. BRISQUE performs well in a range of applications, including the evaluation of image and video compression techniques and medical imaging and surveillance systems.

The BTMQI algorithm evaluates the quality of a tone-mapped image by analyzing three key aspects: information, naturalness, and structure. BTMQI provides a more comprehensive evaluation of tone-mapped image quality and is the latest among these NR-IQMs.

The average scores of 104 images tone mapped using Li, Kwon, Khan, Yang, Hui, Liang, Reinhard, Meylan and TMO$_z$ are reported in Table IV. The lower the scores, the better the quality of the images. TMO$_z$ performed best in all the three NR-IQMs, i.e., PIQE, BRISQUE, and BTMQI with scores of 35.26, 20.62 and 3.78, respectively.

### C. SUBJECTIVE EVALUATION

In the context of the above visual comparison and objective evaluations, the TMO$_z$ algorithm consistently performed well, achieving the best average score across all test images. The above results highlight the effectiveness of TMO$_z$ in producing visually appealing tone mapped images and preserving the original HDR content, such as important details and features. However, subjective evaluation was necessary to evaluate the results using a psychophysical experiment.

The comparison of the objective evaluation showed that the scores of Li, Kwon and Yang TMOs were lower than those of Khan, Liang and Reinhard TMOs. Additionally, the visual comparison results showed that the Liang, Hui and Reinhard TMOs were better than the Li, Kwon, and Yang TMOs. Therefore, the former three TMOs were excluded from the psychophysical experiment.

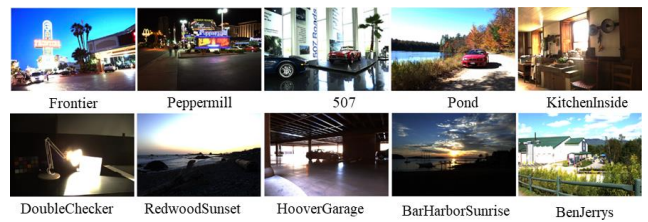

FIGURE 15. Images used to calculate various IQMs scores and rendered for psychophysical experiment, to compare TMO$_z$ with other TMOs.

The subjective evaluation experiment involved comparing TMO$_z$ with other common TMOs, such as Khan, Hui, Liang, Meylan, Reinhard, and Schlick. Reinhard TMO was the most commonly used, while Schlick TMO was noted for producing relatively high-contrast images, as reported in [14]. The Hui TMO produced relatively more saturated images while Meylan TMO produced low-saturated images. To assess the contrast and colorfulness of TMO$_z$, both types of TMOs were employed, i.e., those with high and low contrast and those with high and low colorfulness. This approach allowed for a comprehensive comparison of TMO$_z$ in terms of contrast and color preference.

Ten HDR images from the RIT database were used in a pair comparison experiment. The HDR images included for the rendering purpose are depicted in Figure 15. In three sessions, images were assessed using three terms: Contrast, Colorfulness, and Preference. The display interface was the same as in Figure 8 except for the term used. Provisionally, if the display contrast is higher than the tone mapped image contrast, the images may appear low contrast and vice versa. Since most TMOs do not consider display dynamic range and other factors, images were not assessed in overall preference for contrast and colorfulness. Rather the images were assessed with the following questions per section.

1. Which image has higher contrast?
2. Which image has more colorfulness?
3. Which image do you prefer based on overall preference?

Ten images as depicted in Figure 15 were tone mapped with the seven TMOs and transformed using the Apple Pro XDR display characteristics model as before. The total number of tone mapped images was 70. Since the same display was used in this experiment which was used in brightness compression parameter modeling, similar experimental conditions were applied such as dark room, peak display luminance, distance of the subject from the display, FOV, seating position and comfort level of the subjects. The total number of pairs for comparison was 10 (images) × 7 (TMOs) × (6) / 2 + 30 % repetitions = 273.

TABLE V
Intra- and inter-observer variability of the psychophysical experiment (%age WD).

| Scale | Intra | | | Inter | | |
|---|---|---|---|---|---|---|
| | Mean | Max | SD | Mean | Max | SD |
| Contrast | 25.26 | 47.44 | 10.83 | 34.06 | 53.09 | 6.67 |
| Colorfulness | 15.64 | 26.92 | 7.11 | 21.26 | 24.27 | 1.74 |
| Overall Preference | 20.77 | 29.49 | 4.13 | 30.86 | 34.32 | 1.47 |
| Mean | 20.56 | 34.62 | 7.36 | 28.73 | 37.23 | 3.29 |

Ten subjects performed the experiment in three sessions to assess each of the three image quality measures (contrast, colorfulness, preference). All the subjects were students of the university working in the same college of Optical Science. Specifically, four females and six males performed the experiment with a mean age of 25 and an SD of 2.5. Each subject assessed 273 pairs evaluating a total of 2730 assessments. The average time taken by a subject to perform the entire experiment was approximately 50 minutes.

The validity of experimental data was evaluated by analyzing intra- and inter-observer variability. The percentage of wrong decisions (WDs) for each assessment scale is reported in Table V. If the decision in two assessments differed, it was called a wrong decision (WD) [50]. For the intra-observer variability calculation, individual WDs for individual subjects were considered. The inter-observer variability for each subject was calculated by comparing the data of one subject with each other; then, the mean value was calculated. Table V reports that the mean and maximum values of WDs were higher for the contrast assessment scale than the colorfulness and overall preference. It occurred for both intra- and inter-observer variabilities since the contrast assessment is slightly harder than the colorfulness and overall preference. The impression of colorfulness is marginally easier to understand, so the variabilities of colorfulness were lowest in both cases. The analysis of individual data showed that one subject had intra-observer variability of 47.44 and one subject had 53.09 inter-observer variabilities, resulting in higher SDs. The SDs of intra- and inter-observer variability for colorfulness and overall preference scales were smaller, i.e., 7.11 and 1.74, respectively. Overall, the mean values of the intra- and inter-observer variability were considered reliable for data interpretation and validity of the experiment.

### D. EXPERIMENT RESULTS
The raw data from the experiment was converted into standardized Z-scores using Thurstone law of comparative judgment [51]. Figures 16-18 represent the comparisons of various TMOs for contrast, colorfulness, and overall preference scales, respectively. Figure 16 showed that Schlick and Liang TMOs had the highest contrast scores for each tone mapped image. Meylan TMO had the lowest contrast values for all of the images. Few images tone mapped using Reinhard TMO had higher scores than the mean value while other images had lower scores. The $TMO_z$ achieved moderate contrast for most images and ranked in the middle of these TMOs. Most images of $TMO_z$ had higher contrast than Khan, Hui and Meylan's TMOs, while it was lower than Reinhard, Liang and Schlick images. The mean scores revealed that Schlick TMO ranked at the top, followed by Liang, Reinhard, $TMO_z$, Khan, Hui and Meylan. It has been discussed previously that the contrast of the images is dependent on the display used. Though the images were transformed using the display model, the visual appearance and contrast appearance were affected. The $TMO_z$ achieved moderate contrast ranking compared to the state-of-the-art TMOs and best fits for the displays.

Figure 17 shows the standardized subjective scores of the colorfulness scales for various TMOs. Again, the scores of $TMO_z$ for all images ranked in the middle. The Hui TMO produced the highest colorfulness images. Nine images out of ten were ranked with the highest colorfulness. Khan TMO had similar behavior and all the images had higher colorfulness when compared to the other tone mapped images.

Interestingly, some of the images by Reinhard TMO had more colorfulness than the mean value, while the others had lower scores; however, in all images, it followed Hui and Khan, respectively. Meylan has the lowest colorfulness for 9 images which can be visualized in Figures 11-14. The mean colorfulness subjective scores ranked Hui with the highest colorfulness, followed by Khan, Reinhard, $TMO_z$, Liang, Schlick, and Meylan. The tone mapped image by Meylan TMO had the lowest colorfulness among these TMOs.

In the theory of tone mapping, saturation or colorfulness increases when the contrast is compressed. The saturation of the tone mapped images depends on the contrast compression technique and the resultant contrast of the images. If the tone mapper does not manage saturation or colorfulness, the highly compressed images result in low-contrast and high saturation or colorfulness. Conversely, the images compressed with a lower degree will result in high-contrast images. In turn, the saturation is lower as compared to the former phenomenon. The authors try to reduce the saturation by various algorithms. It can be seen from Figures 12-13 that Schlick and Liang TMOs had the highest contrast, respectively, but had the lowest colorfulness, following the fact explained previously. Hui and Khan TMOs had similar behavior; these TMOs attained lower contrast scores while having the highest colorfulness scores. Meylan TMO ranked lowest in both contrast and colorfulness. Meylan TMO manages the saturation in its algorithm but overly reduces the colorfulness. Reinhard TMO produces slightly higher saturated colors as discussed in [52]. The $TMO_z$ achieved the ranking in the middle. The images produced by $TMO_z$ had colorfulness slightly lower than Reinhard TMO but higher than Schlick, Liang and Meylan TMOs. It conveyed that the images tone mapped using $TMO_z$ had moderate colorfulness. The colorfulness for the $TMO_z$ was adapted from the HDR image using CIECAM16 color adaption equations and display surround conditions. Hence, it can be said be stated that the images produced using $TMO_z$ have more optimal colorfulness for HDR and tone mapping applications.

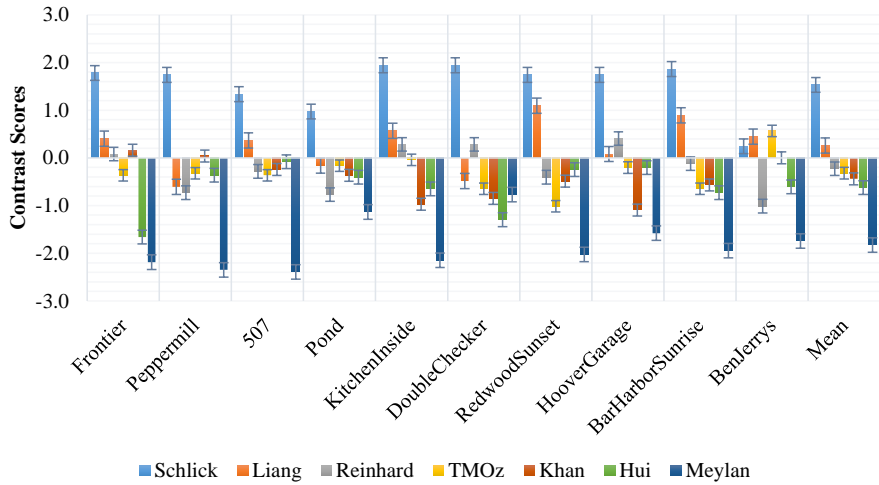

**FIGURE 16.** The comparison TMOs using contrast scale. The higher values of contrast scores indicate high contrast images. TMO$_z$ achieved moderate contrast ranking in the middle.

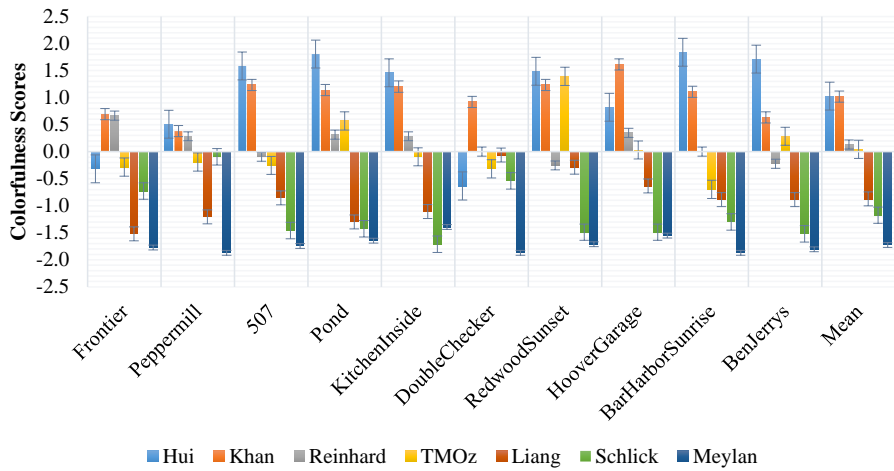

**FIGURE 17.** The comparison TMOs using the colorfulness scale. The higher values of colorfulness scores indicate images with high colorfulness. TMO$_z$ ranked in the middle for the colorfulness scale.

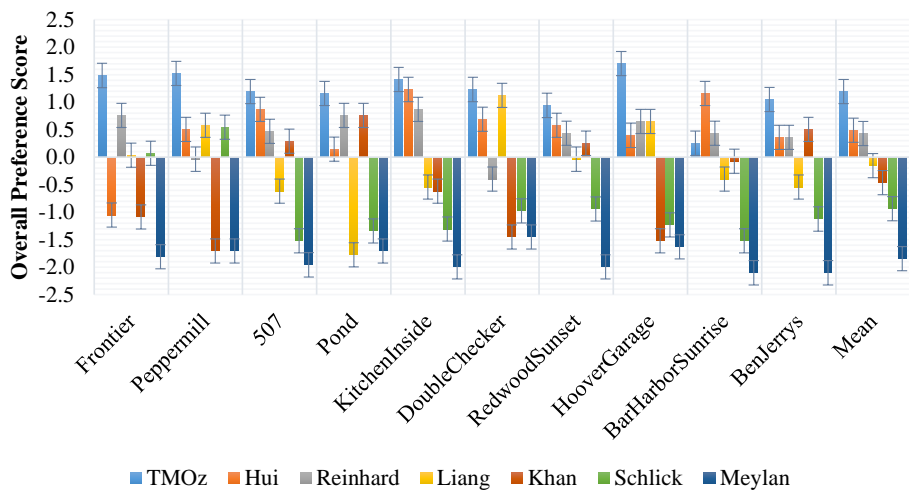

**FIGURE 18.** The overall ranking of the TMOs based on the overall preference scale. TMO$_z$ ranked first due to moderate contrast, moderate colorfulness, more naturalness and better structuredness (as reported using objective metrics).

Lastly, the standardized subjective scores for overall preference are reported in Figure 18. The $TMO_z$ outperformed all of the TMOs in comparison. The colorfulness and contrast of the $TMO_z$ images were moderate; hence it is assertive that the subjects preferred the $TMO_z$ while comparing the other image in the pair. The TMOs ranking for each image varied in overall preference case, except $TMO_z$. Hui TMO performed better than other TMOs in some images while Reinhard or Liang TMOs performed better in others. However, Meylan and Schlick TMOs had the lowest preference scores for all images. It is due to the fact that Meylan TMO had the lowest colorfulness as well as the lowest contrast. Schlick TMO ranked second last in the colorfulness scale yet produced excessively high contrast images. The mean scores ranked the $TMO_z$ on top, Hui TMO on second, followed by Reinhard, Liang, Khan, Schlick, and Meylan TMOs.

*E. SIGNIFICANCE TEST*

The significance of the ranking of the TMOs was tested using a Tukey HSD multi-comparison test combined with Analysis of Variance (ANOVA) [53-55]. The p-value for the current statistics from the ANOVA was $3.0731 \times 10^{-16}$. A p-value of less than 0.05 ($\alpha = 0.05$ at 95% confidence interval) indicates a significant difference in the means of the groups (TMOs). The multi-comparison using the Tukey HSD criterion is depicted in Figure 19.

The plots compare the estimates for different TMOs. Each plot was marked by a symbol, which represented the mean of the TMOs. The confidence interval for each TMO was denoted by a line stretching out from the symbol. The confidence intervals indicated the range of values within which the true mean of the group was likely to lie. The intervals were compared to determine the significance of the differences between the TMO means. If the intervals for two TMOs were disjoint, their means were significantly different. On the other hand, if the intervals overlapped, it suggested that the means were not significantly different. Since the interval of the $TMO_z$ does not overlap with any of the other intervals, it can be asserted that the mean of the $TMO_z$ was significantly different. In turn, the $TMO_z$ performed significantly differently from the other TMOs. As $TMO_z$ was ranked higher in the overall preference case and was significantly different from other TMOs, it can be concluded that $TMO_z$ performed significantly better than the other TMOs.

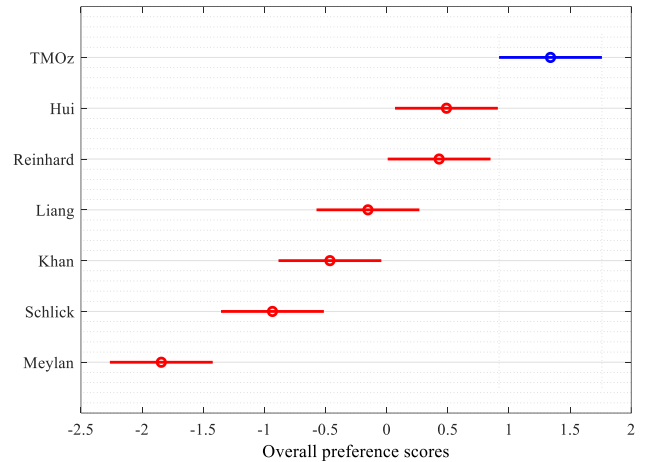

**FIGURE 19.** Multiple Comparison Test of overall preference scale data using ANOVA with Tukey's HSD criterion. $TMO_z$ is significantly different from all other models at a 95% confidence interval.

## V. COMPARISON WITH CNN BASED TMOS

In Figure 20, $TMO_z$ was compared with the recent deep learning based TMOs, i.e., DeepTMO [19] and TMO-NET [20]. The SDR images for these TMOs were directly acquired from the websites of the authors. Both TMOs lead to the excessive saturation problem in all of the images. The hue changes are more prevalent in Figure 20 (a), (b) and (e). Moreover, when the dynamic range is very high such as in Figure 20 (c), (e) and (f), there is a loss of details reproduction in highlights and shadows by CNN based TMOs. However, the images tone mapped using $TMO_z$ have better color and contrast. The images equip well details in highlights and shadows as well.

Moreover, the TMQI score for DeepTMO using [30] database was reported as 0.87. $TMO_z$ achieved a 0.92 score ranking higher than DeepTMO, as reported in Table IV. The TMQI score for TMO-NET using LVZ-HDR dataset [20] was 0.83. We calculated the TMQI scores for $TMO_z$ using the same database; the score was 0.88. It showed $TMO_z$ also outperforms these deep learning based TMOs.

## VI. COMPARISON OF RUNNING TIME

Figure 21 depicts the running time of seven TMOs, including $TMO_z$. The Matlab codes were used for all TMOs. The Matlab 2018b version was used to run these codes on Intel(R), Core (TM) i9-7900X CPU @ 3.30GHz, and 3.31

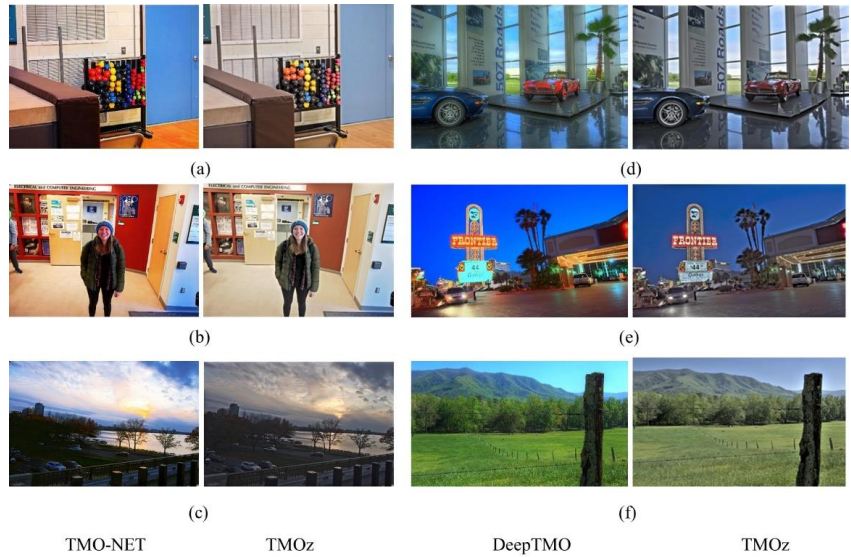

FIGURE 20. Comparison of TMO$_z$ with deep learning based TMOs.

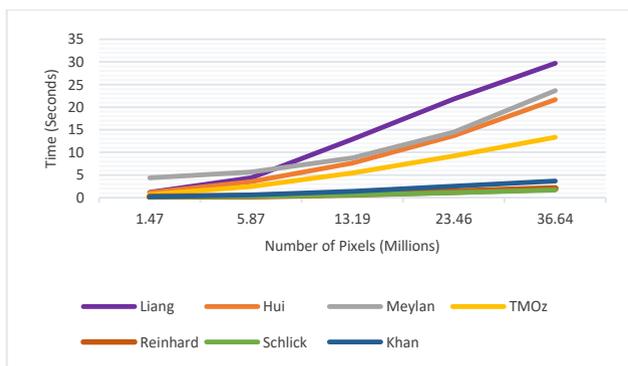

FIGURE 21. Comparison of running time of TMOs.

GHz with 64GB RAM installed. The TMO$_z$ had moderate computation complexity compared to the other TMOs. Most of the time was served in calculating forward and inverse CIECAM16 model parameters. It is obvious from the figure that the computational time of the TMO$_z$ increases less rapidly compared to Hui, Liang and Meylan, which are the most recent TMOs. Hence, the TMO$_z$ has a reasonable running time compared to the other recent methods in the study.

## VII. CONCLUSION

The present work proposed a CIECAM16-based tone mapping model for HDR imaging application. The model called TMO$_z$, compresses and enhances the tone of low- and high- frequency components of the perceptual brightness respectively. This model adopts the colorfulness and hue from the HDR image using CIECAM16 color appearance scales, ensuring more preferred colors in the tone-mapped image. The measures used for the reference-based metric TMIQI and non-reference-based metrics PIQE, NIQE, BRISQUE, and BTMQI were applied to estimate the performance between different TMOs. The results indicate that the TMO$_z$ achieved the highest ranking. A psychophysical experiment was conducted to test the performance of the TMO$_z$ and to compare it with other state-of-the-art TMOs such as Hui, Liang, Reinhard, Khan, Schlick and Meylan.

Each TMO was evaluated against the three sets of visual scores, i.e., preference, contrast, and colorfulness. Again, the TMO$_z$ performed the best to the preference results, but was ranked in the middle for contrast and colorfulness results. TMO$_z$ is also simpler than the other TMOs such as Hui and Liang. Hence, the proposed TMO can be used confidently for the tone mapping of HDR images.

The traditional algorithms only focused on luminance compression except TMO$_z$, which is based on CIECAM16 to take into account the viewing environment such as luminance, color temperature, surround, to produce better and contrast of the images. For the future study, the machine learning techniques will be used to enhance the tone mapping process. By training the model on a large dataset of HDR images and corresponding human preferences, it is possible to create a data-driven approach that can produce faster processor for real-time applications to suit for video streaming or gaming.

## ACKNOWLEDGMENT
The authors thank Dr. Mark Fairchild for providing the HDR image database, i.e., the HDR photographic survey.

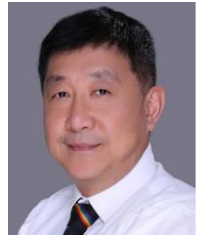

**MING RONNIER LUO** (received his PhD from the Bradford University (UK) on color science in 1986. He is a Chair Professor at the College of Optical Science and Engineering, Zhejiang University (China). He has published over 650 peer reviewed papers in the fields of color science, imaging science and illumination engineering. He is a Fellow of Society of Dyers and Colourists (SDC) and of Imaging Science and Technology (SIST). He has been an active member of International Commission on Illumination (CIE), ex-Vice President, ex-Division Director on Colour and Vision, Technical chairs, members etc. He has received numerous awards, including the recent Judd 2017 Award and the Newton 2020 Award for his contribution in color science research.

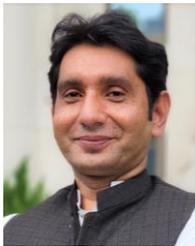

**IMRAN MEHMOOD** received the B.S. and M.S. degrees in electronics from Quaid-i-Azam University, Islamabad, Pakistan. He is currently pursuing the Ph.D. degree with Prof. Ming Ronnier Luo at Color Engineering Lab, College of Optical Science and Engineering, Zhejiang University, Hangzhou, China. He is working on testing and developing generic tone mapping operators in HDR imaging for real-time applications. He has been involved in various projects in color science. He presented his research ideas at international color imaging conferences. His research interests include high dynamic range imaging, tone mapping, color correction, color science, deep learning, etc.

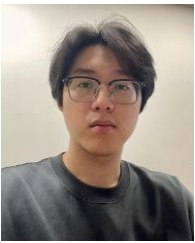

**XINYE SHI** received the Bachelor of Engineering in Optical Science and Engineering from Zhejiang University in 2021. He is currently pursuing the PhD degree in Optical Engineering at Zhejiang University. His research field is color science. His research interests include color appearance models under high dynamic range and tone mapping algorithms.

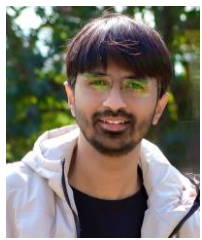

**MUHAMMAD USMAN KHAN** (received his M.Sc. (2011) and M.Phil. (2014) degrees both from Department of Electronics, Quaid-i-Azam University, Islamabad, Pakistan. He is currently doing his Ph.D. in Optical Science and Engineering from Zhejiang University Hangzhou, China.
His research interests include image quality prediction, spatial domain and color domain image quality modeling, HDR imaging etc.